\documentclass[10pt,twocolumn,letterpaper]{article}

\usepackage{cvpr}
\usepackage{times}
\usepackage{epsfig}
\usepackage{graphicx}
\usepackage{amsmath}
\usepackage{amssymb}
\usepackage[nolist]{acronym}
\usepackage{multirow}
\usepackage{subcaption}

\usepackage[pagebackref=true,breaklinks=true,letterpaper=true,colorlinks,bookmarks=false]{hyperref}

\cvprfinalcopy 

\def\cvprPaperID{****} 

\begin{document}

\newacro{lstm}[LSTM]{Long Short-Term Memory}
\newacro{rnn}[RNN]{Recurrent Neural Network}
\newacro{gru}[GRU]{Gated Recurrent Unit}
\newacro{lsta}[LSTA]{Long Short-Term Attention}
\newacro{clstm}[ConvLSTM]{Convolutional Long Short-Term Memory}
\newacro{tsn}[TSN]{Temporal Segment Network}
\newacro{cnn}[CNN]{Convolutional Neural Network}
\newacro{hf-tsn}[HF-TSN]{HF-Temporal Segment Networks}
\newacro{fc}[FC]{Fully Connected}

\title{FBK-HUPBA Submission to the EPIC-Kitchens 2019 \\ Action Recognition Challenge}

\author{Swathikiran Sudhakaran$^1$, Sergio Escalera$^{2,3}$, Oswald Lanz$^{1}$\\[.3cm] 
	$^{1}$Fondazione Bruno Kessler, Trento, Italy\\
    $^{2}$Computer Vision Center, Barcelona, Spain\\
	$^{3}$Universitat de Barcelona, Barcelona, Spain\\
	{\tt\small \{sudhakaran,lanz\}@fbk.eu, \tt\small sergio@maia.ub.es}
}

\maketitle

\begin{abstract}
   In this report we describe the technical details of our submission to the EPIC-Kitchens 2019 action recognition challenge. To participate in the challenge we have developed a number of CNN-LSTA~\cite{lsta} and HF-TSN~\cite{hfnet} variants, and submitted predictions from an ensemble compiled out of these two model families. Our submission, visible on the public leaderboard with team name FBK-HUPBA, achieved a top-1 action recognition accuracy of $35.54\%$ on S1 setting, and $20.25\%$ on S2 setting.
\end{abstract}

\section{Introduction}
Action recognition from videos is one of the most important and ever growing research areas in computer vision. The applications of action recognition range from video surveillance to robotics, human-computer interaction, video indexing and retrieval, \etc. The availability of graphics processing units (GPUs) and large scale datasets have resulted in the development of several data-driven techniques for action recognition via deep learning. EPIC-Kitchens dataset~\cite{damen2018scaling} consists of egocentric videos. Recognition of actions classes in this dataset is challenged by the need for a fine-grained discrimination of small objects and their manipulation.

\begin{figure*}
    \centering
    \begin{subfigure}[b]{0.45\textwidth}
    \includegraphics[scale=0.4]{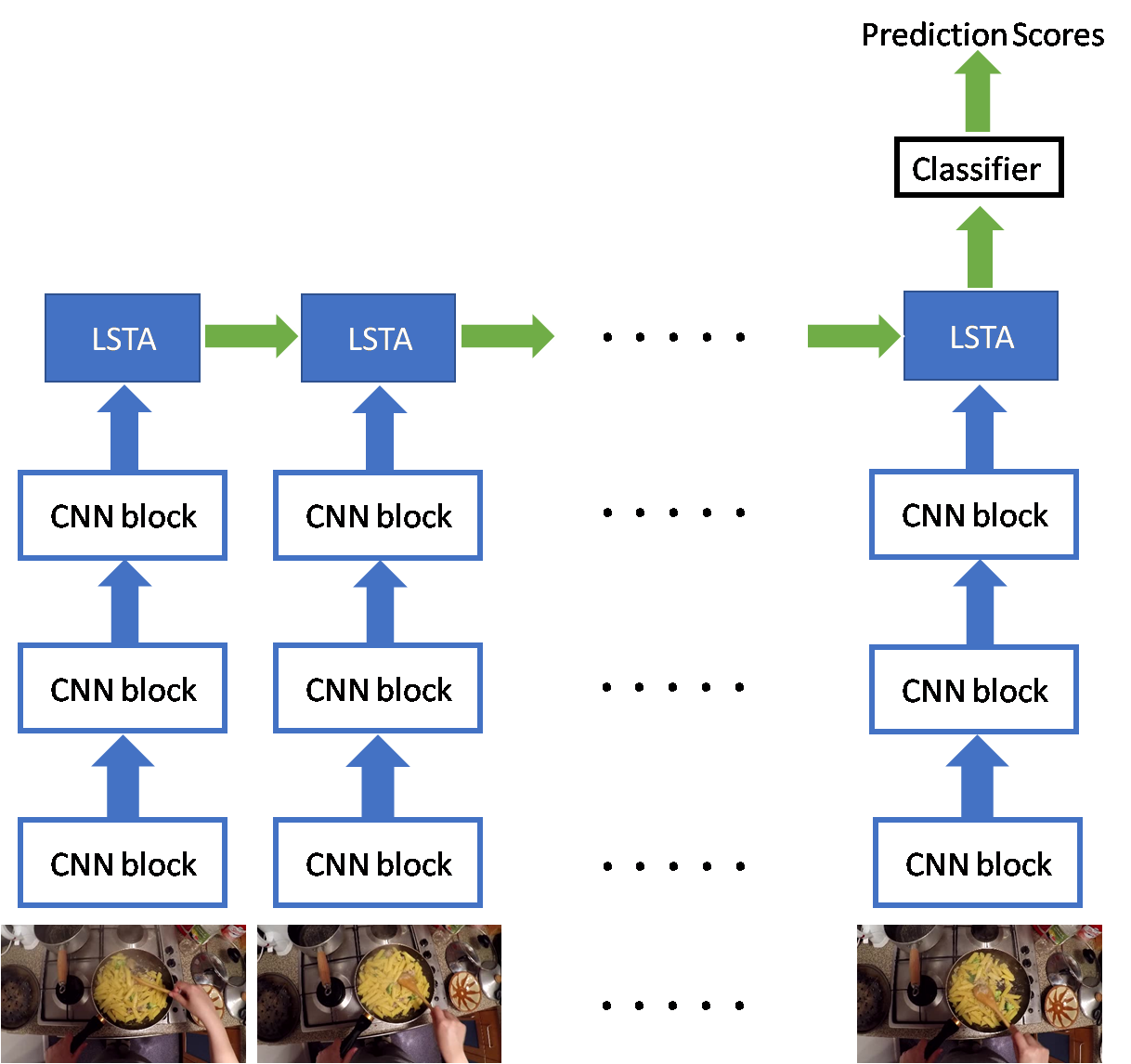}
    \caption{CNN-LSTA}
    \label{fig:lsta}
    \end{subfigure}\hfill
    \begin{subfigure}[b]{0.45\textwidth}
    \includegraphics[scale=0.4]{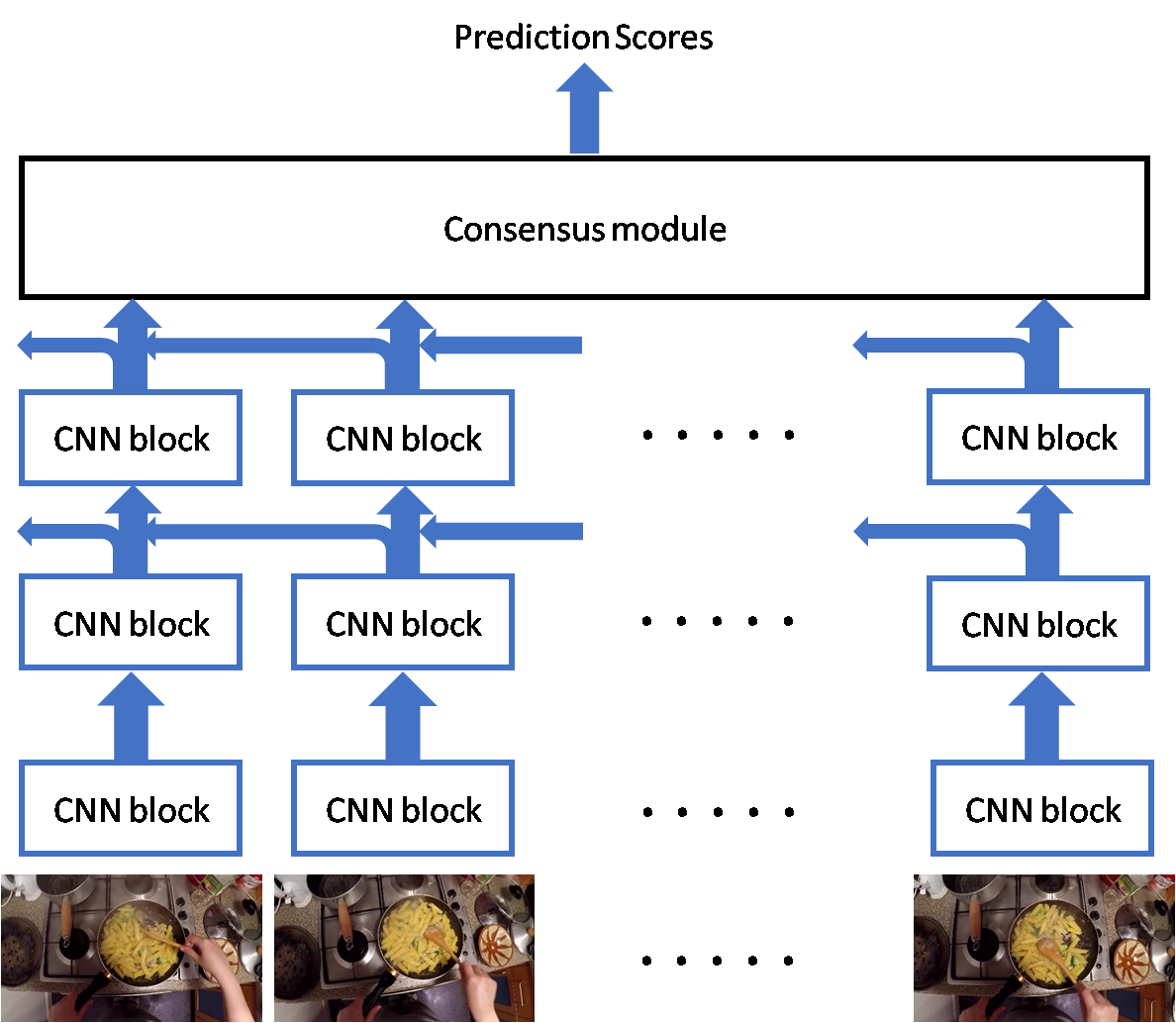}
    \caption{HF-TSN}
    \label{fig:hf_tsn}
    \end{subfigure}
    \caption{Block diagram illustrating the two model families used for generating the action recognition scores. The first model in Fig.~\ref{fig:lsta} uses a \ac{lsta} module to aggregate the frame level features obtained from a \ac{cnn} backbone. This is equivalent to late fusion of the frame level features. In the second method, Fig.~\ref{fig:hf_tsn}, features from adjacent frames are combined as the inputs move across the \ac{cnn} layers, followed by a late fusion of the features obtained at the final layer of the \ac{cnn}. Thus, the two considered approaches provide complementary ways to aggregate frame level features.
}
    \label{fig:block_diagram}
\end{figure*}

For our participation to the challenge we considered two different approaches with complementary feature encoding perspective for classifying action categories:
\begin{itemize}
    \item CNN-LSTA~\cite{lsta}: late (and shallow) aggregation of frame level features with a variant of LSTM;
    \item HF-Nets~\cite{hfnet}: early (and deep) aggregation of frame level features using a temporal gating mechanism.
\end{itemize}
Fig.~\ref{fig:block_diagram} shows block diagrams of the two different approaches, both of them developed by the FBK-HUPBA team. For a detailed presentation of the two baseline methods we refer the reader to the original papers~\cite{lsta,hfnet}. 

To participate in the challenge we have developed variants of both CNN-LSTA and HF-TSN baselines. We have changed backbone CNNs, enriched the aggregation scheme of LSTA, implemented a structured prediction, and differentiated training strategies. We finally compiled an ensemble out of this pool of trained models. Our submission visible on the public leaderboard was obtained by averaging classification scores from ensemble members.

\section{CNN-LSTA and variants}
\label{sec:lsta_variants}

Our first family of models is CNN-RNN structured. The RNN is a Long Short-Term Attention (LSTA) recurrent unit~\cite{lsta}. In brief, LSTA extends LSTM with built-in attention and a revised output gating. Attention is introduced to promote discriminative features in the memory updating. This is done by applying a spatial weight map to the input. Output pooling provides more flexibility in localizing and propagating the active memory components. 

We have modified CNN-LSTA baseline as follows:
\begin{itemize}
    \item Backbone: we used ResNet-34, ResNet-50, InceptionV3;
    \item Pre-training: we utilized pretrained models on ImageNet and Kinetics;
    \item Aggregation: we used LSTA internal memory as aggregated descriptor for classification as in~\cite{lsta}, but we also aggregated the sequence of output states using GRU and concatenated its final memory state with that of the LSTA for classification.
\end{itemize}

For the variant with \ac{gru}, the output state of LSTA during each time step is spatial average pooled and applied to two \acp{gru}. The output states of the \acp{gru} after encoding all the video frames are then concatenated to predict the \verb|verb|, \verb|noun| and \verb|action| classes. The scores generated from \ac{gru} and \ac{lsta} are then averaged to obtain the corresponding class scores. Structured prediction is detailed in Sec.~\ref{sec:structred_prediction}.

\section{HF-TSN and variants}
\label{sec:hf-tsn_variants}

Our second pool are TSN models with hierarchical feature aggregation~\cite{hfnet}. In HF-TSN, features from adjacent frames of a video interact with each other as the features are being passed along the layers of a \ac{cnn}. The interactions are learned and comprises of either differencing or averaging operation, or a mixture of them, via a convolutional layer. The features corresponding to each spatio-temporal receptive field, obtained at the final layer of the \ac{cnn}, are applied to a linear layer and averaged to obtain the action class score. The consensus module in Fig.~\ref{fig:hf_tsn} represents the linear layer followed by averaging operation.

We have modified HF-TSN baseline as follows:
\begin{itemize}
    \item Backbone: we used ResNet-50 and BNInception.
\end{itemize}

For the model with ResNet-50, HF blocks are applied at the input of each of the ResNet-50 blocks. Thus a total of 16 HF blocks are present in this variant as opposed to the 10 present in the model with BNInception.

\section{Structured prediction}
\label{sec:structred_prediction}
The labels provided with the dataset are in the form of \verb|verb| and \verb|noun| pairs. An action is defined by the combination of such \verb|verb|-\verb|noun| pairs. So the network should be able to either correctly predict both the \verb|verb| and \verb|noun| classes in order to combine them into an \verb|action| class, or directly predict the \verb|action| class from which \verb|verb| and \verb|noun| classes can be derived. We trained all the networks as a multi-task classification problem predicting \verb|verb|, \verb|noun| and \verb|action| classes. We generated \verb|action| classes from the combination of \verb|verb| and \verb|noun| labels present in the dataset. It is important to note that not all combinations of \verb|verb|-\verb|noun| pairs are valid, such as, \verb|take|-\verb|fridge|, \verb|open|-\verb|carrot|, \verb|cut|-\verb|salt| are unfeasible.

In order to model such inter-dependencies among the \verb|verb| and \verb|noun| classes, we apply the \verb|action| prediction scores as an instance-specific bias term to the \verb|verb| and \verb|noun| classifiers. For this, the \verb|action| scores are applied through two linear layers each to map to the number of \verb|verb| and \verb|noun| classes. The result is then applied to the output of the corresponding classifier (\verb|verb| and \verb|noun|). This allows the network to learn the dependencies between the \verb|verb| and \verb|noun| classes and prevent it from making unfeasible predictions consisting of implausible \verb|verb| and \verb|noun| combinations. The drawback of this approach is that we are bound to predict \verb|action| classes observed during training.

\section{Cross-modal fusion}
\label{sec:cross_modal}
For \ac{lsta} model, we also implement a two stream model with cross-modal fusion. We follow the approach proposed in~\cite{lsta} for the two stream implementation. 

The \ac{lsta} model with ResNet-34 \ac{cnn} is used as the appearance stream. For the motion stream, we first trained a ResNet-34 \ac{cnn} pre-trained on ImageNet for predicting \verb|verb| classes followed by a separate training stage for predicting \verb|verb|, \verb|noun| and \verb|acion| classes. A stack of optical flow images corresponding to 5 consecutive frames is used as the input to the network. The first convolutional layer of the network is modified to accept an input image with 10 channels and the weights are initialized by averaging the weights from the three channels of the original network.

Once the appearance and motion stream networks are trained separately, we combine them using cross-modal fusion and fine-tune the parameters. In order to perform cross-modal fusion, we first add a \ac{clstm} layer, with a hidden size of 512, after the \verb|conv5_3| layer of the motion stream. Then the outputs corresponding to each of the frames from the \verb|conv5_3| layer of the appearance stream are combined using a 3D convolution layer, which is applied as bias to the gates of the \ac{clstm} layer. Similarly, the output from the \verb|conv5_3| layer of the motion stream is applied as bias to the gates of the \ac{lsta} layer present in the appearance stream. Finally, the classification scores from the two individual streams are averaged to obtain the final prediction score of the video.

\section{Training details}

In this section we provide details on the training protocol. 
We did not use a held-out validation set for hyperparameter search or model validation.

\subsection{CNN-LSTA variants}
We used the same training strategy presented in~\ac{lsta}~\cite{lsta}, \ie the networks are trained in two stages. In the first stage, the classification layers and \ac{lsta} layer (and the \acp{gru} in the case of variant 3) are trained for 200 epochs starting with a learning rate of 0.001 which is decayed by a factor of 0.1 after 25, 75 and 150 epochs. During stage 2, the \verb|conv5_x| layer in the case of ResNet family of \acp{cnn} or \verb|Mixed_7x| layers in the case of InceptionV3, are trained in addition to the layers trained during stage 1. Stage 2 training is done for 150 epochs with an initial learning rate of 0.0001 which is decayed by a factor of 0.1 after 25 and 75 epochs. A dropout of 0.7 is used to avoid overfitting. ADAM algorithm is used for the optimization of the parameters with a batch size of 32 during training. 20 frames selected uniformly across time are used as the input during both training and evaluation stages. We use random scaling and horizontal flipping as data augmentation techniques during training and during evaluation, we average the scores obtained from five crops (four corner crops and the center crop) and their horizontally flipped versions. In all the models, \ac{lsta} and \ac{gru} with a memory size of 512 is used. The dimension of the input to the ResNet models is $224\times224$ and for InceptionV3 is $299\times299$.

\subsection{HF-TSN variants}

The models are trained for 120 epochs with an initial learning rate of 0.01 that is decayed by a factor of 0.1 after 50 and 100 epochs. We used a batch size of 32 and dropout of 0.5 to prevent overfitting. Stochastic Gradient Descent (SGD) is used as the optimization algorithm. Spatial scaling and random horizontal flipping with temporal jittering is used as data augmentation techniques. During evaluation, 10 image crops are generated from each frame using cropping and horizontal flipping and their average of scores is used for predicting the action class of the video. 16 frames are sampled from each video during training and inference. The input image dimension is set as $224\times224$.

\subsection{Two-stream variants}

For the flow stream, the network is trained for 700 epochs, for \verb|verb| classification, with an initial learning rate of 0.01 which is reduced by 0.5 after 75, 150, 250 and 500 epochs. This acts as a pre-training for the network. After this, we train the network for action classification with the same structured prediction technique explained in~\ref{sec:structred_prediction}. We also apply spatial attention to the features at the output of the \verb|conv5_3| layer. We follow the idea proposed in~\cite{egornn} for applying spatial attention to the motion features. During this stage, the network is trained for 500 epochs with a learning rate of 0.01. The learning rate is decayed after 50 and 100 epochs by 0.5. SGD algorithm is used for optimizing the parameter updates of the network in both stages.
 
For the two stream model, the networks are finetuned for 100 epochs with a learning rate of 0.01 using ADAM algorithm. Learning rate is reduced by a factor of 0.99 after each epoch. We finetune the classification layers, \ac{lsta}, \ac{clstm} and \verb|conv5_x| layers of the two networks in this stage.

\begin{table*}[th]\small
	\centering
	\begin{tabular}{c|l|l|c|c|c|c|c|c|c|c|c|c|c|c}
		\hline
		& \multirow{2}{*}{Method} & \multirow{2}{*}{Backbone} & \multicolumn{3}{c|}{Top-1 Accuracy (\%)} & \multicolumn{3}{c|}{Top-5 Accuracy (\%)} & \multicolumn{3}{c|}{Precision (\%)} & \multicolumn{3}{c}{Recall (\%)} \\
		\cline{4-15}
		& & & \parbox{0.7cm}{Verb} & \parbox{0.7cm}{Noun} & \parbox{0.8cm}{Action} & \parbox{0.7cm}{Verb} & \parbox{0.7cm}{Noun} & \parbox{0.8cm}{Action} & \parbox{0.7cm}{Verb} & \parbox{0.7cm}{Noun} & \parbox{0.8cm}{Action} & \parbox{0.7cm}{Verb} & \parbox{0.7cm}{Noun} & \parbox{0.8cm}{Action}\\
		\hline \hline
		\multirow{11}{*}{\rotatebox[origin=t]{90}{S1}}
		& \multirow{4}{*}{LSTA} & Res-34 & 58.25 & 38.93 & 30.16 & 86.57 & 62.96 & 50.16 & 44.09 & 36.30 & 16.54 & 37.32 & 36.52 & 19.00\\ \cline{3-15}
		&  & Res-50 & 57.81 & 37.84 & 29.54 & 86.14 & 63.63 & 49.82 & 52.76 & 34.77 & 16.35 & 33.94 & 34.46 & 18.05\\ \cline{3-15}
		&  & Res-50$^\dagger$ & 57.69 & 39.36 & 29.79 & 86.77 & 64.46 & 50.52 & 50.83 & 36.49 & 17.54 & 33.68 & 35.70 & 17.38\\ \cline{3-15}
		&  & IncV3 & 57.28 & 39.32 & 29.35 & 86.43 & 64.32 & 50.18 & 54.77 & 36.08 & 14.51 & 34.29 & 35.64 & 16.65\\ \cline{2-15}
		& \multirow{3}{*}{LSTA-\ac{gru}} & Res-50$^{*}$ & 57.30 & 37.59 & 29.17 & 85.88 & 62.97 & 49.24 & 49.32 & 34.79 & 16.81 & 34.84 & 34.33 & 18.40\\ \cline{3-15}
		&  & Res-34$^{**}$ & 60.61 & 40.84 & 32.04 & 87.71 & 65.93 & 52.75 & 53.62 & 37.29 & 18.74 & 36.75 & 37.30 & 19.76\\ \cline{3-15}
		&  & Res-34$^{***}$ & 61.31 & 40.93 & 32.14 & 87.47 & 65.28 & 52.60 & 50.93 & 38.23 & 19.59 & 37.90 & 37.47 & 20.36\\ \cline{2-15}
		& LSTA-2S & Res-34 & 62.12 & 40.41 & 32.60 & 87.95 & 64.47 & 52.85 & 52.70 & 39.66 & 15.95 & 36.34 & 36.88 & 18.61\\
		\cline{2-15}
		& \multirow{2}{*}{HF-TSN} & BNInc & 57.57 & 39.90 & 28.09 & 87.83 & 65.37 & 48.63 & 49.12 & 35.83 & 11.38 & \textbf{39.37} & 37.04 & 13.84\\
		\cline{3-15}
		&  & Res-50 & 56.69 & 40.70 & 29.38 & 86.47 & 63.91 & 49.36 & 41.88 & 37.91 & 10.70 & 37.86 & 38.52 & 13.58\\ \cline{2-15}
		& \multicolumn{2}{c|}{\textbf{Ensemble}}  & \textbf{63.34} & \textbf{44.75} & \textbf{35.54} & \textbf{89.01} & \textbf{69.88} & \textbf{57.18} & \textbf{63.21} & \textbf{42.26} & \textbf{19.76} & 37.77 & \textbf{41.28} & \textbf{21.19} \\
	 \hline
        \multirow{11}{*}{\rotatebox[origin=t]{90}{S2}} 
        & \multirow{4}{*}{LSTA} & Res-34 & 45.51 & 23.46 & 15.88 & 75.25 & 43.16 & 30.01 & 26.19 & 17.58 & 8.44 & 20.80 & 19.67 & 11.29\\ \cline{3-15}
        &  & Res-50 & 44.38 & 22.53 & 15.98 & 74.29 & 43.02 & 30.42 & 23.36 & 17.69 & 7.31 & 17.39 & 17.92 & 10.29\\ \cline{3-15}
        &  & Res-50$^\dagger$ & 43.53 & 22.98 & 16.25 & 74.70 & 44.66 & 30.01 & 22.05 & 15.70 & 7.81 & 15.73 & 17.62 & 10.83\\ \cline{3-15}
        &  & IncV3 & 44.66 & 23.76 & 17.31 & 75.35 & 47.97 & 32.64 & 24.69 & 17.80 & 7.70 & 16.10 & 19.38 & 11.19\\ \cline{2-15}
        & \multirow{3}{*}{LSTA-\ac{gru}} & Res-50$^{*}$ & 43.94 & 22.16 & 15.94 & 73.61 & 42.47 & 29.70 & 23.20 & 17.84 & 8.24 & 17.04 & 17.71 & 10.27\\ \cline{3-15}
        &  & Res-34$^{**}$ & 45.37 & 23.49 & 16.59 & 74.74 & 45.24 & 31.17 & 30.04 & 16.05 & 7.51 & 16.38 & 17.93 & 10.23\\ \cline{3-15}
        &  & Res-34$^{***}$ & 44.90 & 22.60 & 16.25 & 74.80 & 44.62 & 31.14 & 32.62 & 16.45 & 7.87 & 17.99 & 19.41 & 10.53\\ \cline{2-15}
 		& LSTA-2S & Res-34 & 48.89 & 24.27 & 18.71 & \textbf{77.88} & 46.06 & 33.77 & 27.12 & 20.12 & \textbf{9.29} & \textbf{22.59} & 18.91 & 12.91\\ 
 		\cline{2-15}
		& \multirow{2}{*}{HF-TSN} & BNInc & 42.40 & 25.23 & 16.93 & 75.76 & 48.96 & 33.32 & 24.25 & 20.48 & 6.29 & 15.77 & \textbf{21.96} & 10.05 \\ 
		\cline{3-15}
		&  & Res-50 & 45.48 & 24.55 & 17.38 & 75.32 & 46.91 & 33.32 & 29.44 & \textbf{22.94} & 7.44 & 19.11 & 21.05 & 10.68\\ \cline{2-15}
		& \multicolumn{2}{c|}{\textbf{Ensemble}}  & \textbf{49.37} & \textbf{27.11} & \textbf{20.25} & 77.50 & \textbf{51.96} & \textbf{37.56} & \textbf{31.09} & 21.06 & 9.18 & 18.73 & 21.88 & \textbf{14.23}\\
		\hline 
	\end{tabular}
	\caption{Comparison of recognition accuracies with state-of-the-art in EPIC-KITCHENS dataset. $^\dagger$: Kinetics pre-trained; $^{*}$: finetuned layers- GRU; $^{**}$: finetuned layers- GRU+LSTA; $^{***}$- finetuned layers- GRU+LSTA+\texttt{Conv5\symbol{95}3}}
	\label{tab:epic_kitchens}
\end{table*}

\section{Results}
\label{sec:results}

The recognition accuracy obtained for each of the selected models and their ensemble are listed in Tab.~\ref{tab:epic_kitchens}. Since no validation set is provided with the dataset, we choose models for ensembling based on their design variability. Each selected model has been submitted for  evaluation on the test server. Model ensembling is done by averaging the prediction scores obtained from individual models. We participated to the challenge with the ensemble.

The best performance obtained for S1 using RGB frames is by the \ac{lsta} model with \acp{gru} encoding the output state of \ac{lsta}. The model resulted in a recognition accuracy of $32.14\%$. Using the cross-modal fusion technique explained in Sec.~\ref{sec:cross_modal}, the recognition accuracy improved by $2\%$ ($30.16$ vs $32.60$). 
By combining the \ac{lsta}-2S and HF-TSN-BNInception models, an improvement of $1\%$ is obtained. With an ensemble of all the models, the action recognition accuracy is further improved by $2\%$.

In S2 setting, the best performance using RGB frames as input was obtained by HF-TSN model with ResNet-50 backbone ($17.38\%$). A gain of about $3\%$ is obtained using cross-modal fusion over the \ac{lsta} model. A gain of $2\%$ is obtained from an ensemble of \ac{lsta}-2S and HF-TSN-BNInception. The ensemble of all the models resulted in an accuracy of $20.25\%$. This proves that the selection of models based on the difference in training settings and temporal encoding techniques was beneficial.

\section{Conclusions}
We described the details of the two model families and their variants  we ensembled for our submission to the action recognition task of the EPIC-Kitchens CVPR 2019 challenge. The recognition accuracy obtained shows that the two model families perform complementary temporal encoding of features. With an ensemble of the proposed methods, our entry to the challenge achieved the score of $35.54\%$ on S1 setting, and $20.25\%$ on S2 setting.

{\small
\bibliographystyle{ieee_fullname}
\bibliography{epic_bib}
}

\end{document}